\documentclass[conference]{IEEEtran}
\IEEEoverridecommandlockouts
\usepackage{cite}
\usepackage{amsmath,amssymb,amsfonts}
\usepackage{algorithmicx}
\usepackage{graphicx}
\usepackage{textcomp}
\usepackage{xcolor}
\usepackage{scalerel}
\usepackage{tikz}
\usepackage{xurl}
\usetikzlibrary{svg.path}

\newcommand{\orcidicon}[1]{%
\href{https://orcid.org/#1}{%
\includegraphics[width=1.0em]{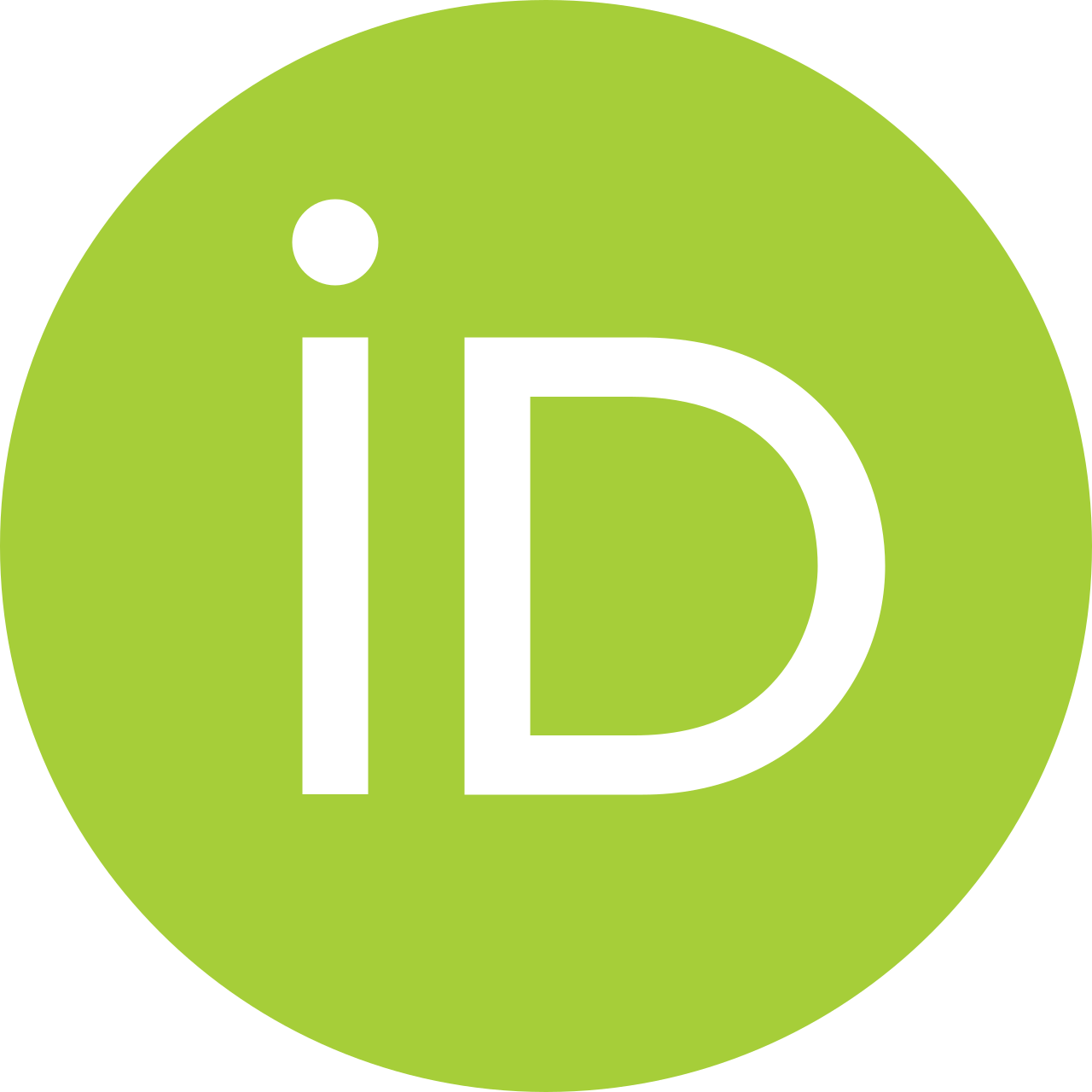}%
}%
}

\usepackage{hyperref} %<--- Load after everything else
\def\BibTeX{{\rm B\kern-.05em{\sc i\kern-.025em b}\kern-.08em
    T\kern-.1667em\lower.7ex\hbox{E}\kern-.125emX}}
\begin{document}

\title{Clearing the Underbrush:\\AI-Enhanced RF Interference Suppression
\thanks{Research was sponsored by the Department of the Air Force Artificial Intelligence Accelerator and was accomplished under Cooperative Agreement Number FA8750-19-2-1000. The views and conclusions contained in this document are those of the authors and should not be interpreted as representing the official policies, either expressed or implied, of the Department of the Air Force or the U.S. Government. The U.S. Government is authorized to reproduce and distribute reprints for Government purposes notwithstanding any copyright notation herein.}
}

\author{
    Rahul Jain \orcidicon{0009-0009-3723-5720}, Pierre Trepagnier \orcidicon{0000-0003-2869-1504}, Rick Gentile \orcidicon{0009-0004-7758-1947}, Joey Botero \orcidicon{0000-0001-9848-8611}, Alexia Schulz \orcidicon{0000-0002-4143-8792}\\
    \textit{MIT Lincoln Laboratory, Lexington, MA 02421, USA}\\
    \small {\{Rahul.Jain, ptrepagnier, Richard.Gentile, Joey.Botero, Alexia.Schulz\}}@ll.mit.edu\\
}

\maketitle

\begin{abstract}

AI-based structured interference rejection has grown more popular because deep learning approaches can outperform traditional methods by jointly considering the signal of interest (SOI) and the signal mixture (SOI plus interference). This work builds on a previous AI-enabled approach utilizing autoregressive transformer-based models by adding a Finite Scalar Quantization (FSQ) tokenizer layer which aims to improve the interference rejection performance while keeping overall latency to a minimum. Additionally, we experiment with other inference optimization techniques with the goal of speeding up inference without much accuracy loss. We explore this space with an experiment where the SOI is a digitally modulated radio frequency (RF) signal and the structured interference is a digital television signal, an extremely common type of Orthogonal Frequency-Division Multiplexing (OFDM) transmission. Our results achieve low latency and increased interference rejection over traditional techniques and prior work with other AI-enabled methods. We demonstrate the benefits of the AI-enabled approaches via audio metrics such as Perceptual Evaluation of Speech Quality (PESQ). Additionally, we explore a variety of applications and detail how our interference rejection algorithm may be used in operationally-relevant scenarios.

\end{abstract}


\section{Introduction}

\begin{figure}[htbp]
\centerline{\includegraphics[width=0.7\columnwidth]{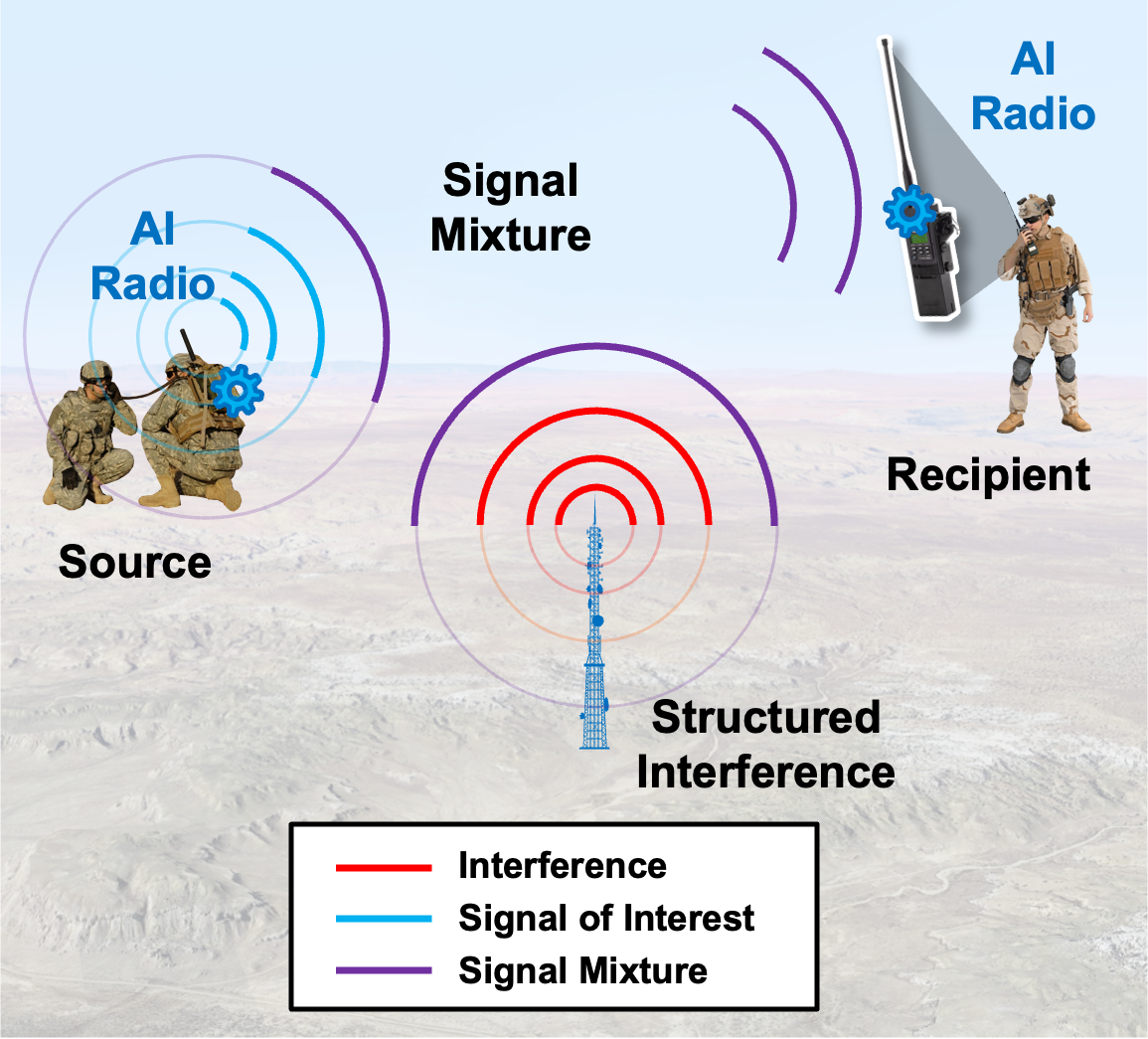}}\caption{ A real-time resilient communications use case. The blue gear box represents the AI model doing the processing on the receive end of the communications link.}
\label{fig:intro}
\end{figure}

This work reports on AI-enhanced RF interference suppression research undertaken as part of the Department of the Air Force-MIT AI Accelerator (AIA), a collaboration involving the Air Force, MIT, and MIT Lincoln Laboratory. Many of the theoretical advances on the network architectures were originated by our partner researchers at MIT \cite{LifarRFTransformer}. The work presented is primarily focused on operationalizing these advances for future deployment. Our research group applies AI to national security challenges, particularly at the intersection of cyber and physical systems. In this case, radio-frequency (RF) communication systems are of interest. In this paper, we examine the use of AI to enhance RF interference rejection on software-defined radios at the tactical edge. Figure \ref{fig:intro} shows an example of a real-time resilient communications use case where the goal is to recover a signal of interest (SOI) from the signal mixture.

 Systems at the tactical edge present challenges to AI deployment beyond those found in academic or enterprise settings. AI deployment at the tactical edge is subject to severe size, weight, and power (SWaP) constraints, yet it must still perform inference fast enough for real-time or near-real-time processing. Because communications with enterprise-scale computing capabilities could be degraded or nonexistent in operational situations, AI-enabled systems need to be trained to handle interferers with zero- or few-shot learning. However, the case for AI-enabled interference mitigation is still compelling. With the exception of simple repetitive interferers, interference mitigation has previously focused on enhancing the SOI (e.g., with a matched filter) rather than suppressing the interferer. However, the wideband interferers that we consider are rich in structure. Deep-learning-based models, trained on the signal mixture as well as the SOI, can pick out features of the SOI and interference, enabling the interference to be removed and the SOI to be recovered. While traditional techniques can be used when details of the SOI and interferer are known, our approach does not rely on detailed, design-level knowledge of the interfering signal or the propagation conditions. We will demonstrate interference mitigation using AI on digitally modulated signals that are commonly found in tactical radios. However, as we will discuss, these same techniques can be applied to a broad range of commercial applications including those not confined to the tactical edge. 

\section{Background}

\subsection{Challenges in the Wireless RF Domain}

 Modern wireless communication systems have to operate in challenging conditions and environments due to the presence of intentional and unintentional interference sources. Robust recovery of SOIs requires operations over a range of signal-to-interference-plus-noise ratios (SINR). Transmitted signals reach the receiver after traveling over multiple propagation paths which introduce varied amplitudes, delays, and phases that produce constructive and/or destructive interference at the receiver. These conditions can directly lead to bit errors in the processed signals. These factors all pose challenges to meeting the associated link budget requirements needed to recover a SOI. The RF spectrum is also crowded, making it difficult to find interference-free frequency bands. New frequency bands are consumed as fast as they become available by commercial and military applications. Waveform bandwidths also continue to grow as commercial software-defined radios with high-speed transceivers rapidly evolve. In military applications, the spectrum is further reduced with the presence of evolving adversary wideband multifunction RF systems. 

 An example of a common multicarrier modulation technique used in commercial and military systems is Orthogonal Frequency-Division Multiplexing (OFDM). It is a ubiquitous communications waveform because it enables high data rates by using multiple lower-rate streams modulated in parallel. Each stream is modulated with a separate subcarrier. The subcarrier frequencies are orthogonal, which avoids the spectral overlap that causes inter-carrier interference (ICI). Having multiple narrowband subcarriers rather than one wideband signal helps to improve the effects of signal fading because the subcarriers are on different frequencies. In addition, because the subcarriers are spaced closely together, a large number of subcarriers can fit within an allocated channel bandwidth. The total bandwidth of an OFDM signal is roughly equal to the sum of the individual subcarrier bandwidths, which makes it much wider than a typical single-carrier SOI. The subcarrier modulation and spacing provide structure in the time-frequency domains which an AI-enabled approach can learn from processing large training datasets. This is true even when the SOI is digitally modulated, as is the case in our experiment.

 While spatial signal processing techniques that leverage phased array antennas can be used to improve SOI gain by removing the effects of interference sources with algorithms like beamforming and adaptive nulling \cite{11123603}, the techniques described in this paper are more applicable to SWaP-constrained systems where only a single antenna element is present.

\section{Approach}

 The approach presented in this paper builds upon similar methodologies \cite{JainICADpaper, LifarRFTransformer} by advancing from traditional to AI-based techniques for RF interference rejection. Some common traditional techniques include matched filtering, Linear Minimum Mean Squared Error (LMMSE), and successive interference cancellation (SIC) \cite{kay1993fundamentals, moshavi1996multiuser}. However, these approaches are either suboptimal in the presence of structured interference, too slow due to the large computational load, or assume detailed knowledge of the channel and/or interference source. In contrast, the AI-based techniques, including RF WaveNet, RF Transformer, and RF Transformer Decoder, are effective for structured interference rejection, and depending on the deployment environment may be fast enough for operational use. Our goal is to explore variations of the transformer-based models to improve signal quality and/or inference time.

\subsection{RF FSQ Tokenizer}

 Many applications require representing high-dimensional or continuous data compactly, which is especially critical in the RF domain, where high sample rates produce large data volumes that challenge efficient model learning. To address this, we used a tokenizer to represent continuous RF data sequences as a finite set of discrete values, allowing our models to learn a compressed representation rather than the raw continuous signal. There are multiple ways to implement this including vector quantization, finite scalar quantization, and product quantization. Lifar et al. \cite{LifarRFTransformer} have shown encouraging results with the finite scalar quantization method, so that will be the focus of this discussion. 

\begin{figure*}[htbp]
\centerline{\includegraphics[width=0.8\textwidth]{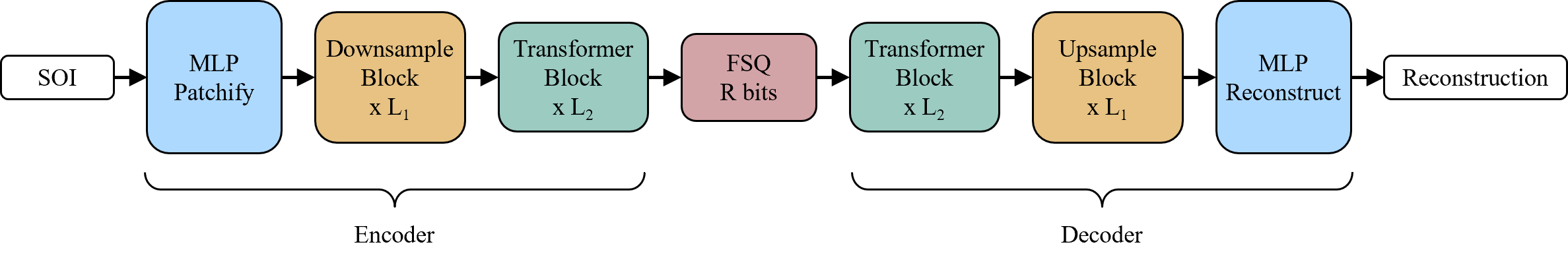}}\caption{ Adapted from \cite{LifarRFTransformer}. Architecture diagram for the RF Finite Scalar Quantization (FSQ) tokenizer.}
\label{fig:fsq_tokenizer}
\end{figure*}

 Finite Scalar Quantization (FSQ) is a technique that maps continuous data points $z$ to a finite set of discrete values by partitioning the input space into regions and assigning each region a representative value $\hat{z} = \text{round}(f(z))$, where $f$ is a bounding function \cite{mentzer2023finitescalarquantizationvqvae, LifarRFTransformer}. The resulting $\hat{z}$ is the compact representation of the original data and is commonly referred to as a ``token''. Figure \ref{fig:fsq_tokenizer} shows the RF FSQ tokenizer architecture diagram, which includes a combination of a Multi-Layer Perceptron (MLP), downsample, upsample, and transformer blocks \cite{LifarRFTransformer}. The tokenizer was trained with a mean squared error reconstruction loss objective, nearly causal configuration, and with a signal length that matched the transformer-based models' signal length, though other configurations are possible.

\subsection{RF Transformer using the RF FSQ Tokenizer}

\begin{figure}[htbp]
\centerline{\includegraphics[width=0.7\columnwidth]{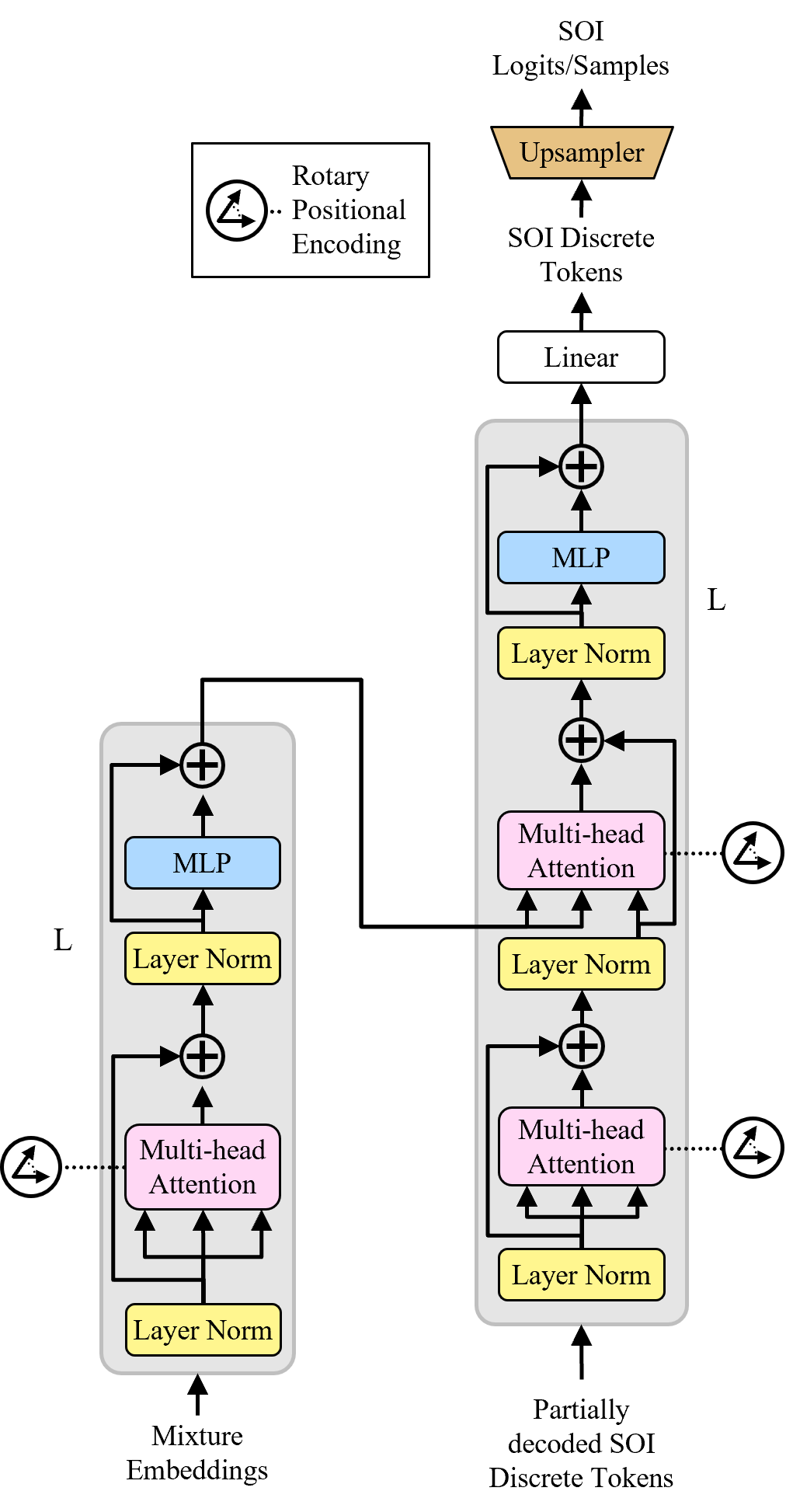}}\caption{ Adapted from \cite{LifarRFTransformer, JainICADpaper}. Architecture diagram for the RF Transformer using the RF FSQ tokenizer.}
\label{fig:transformers}
\end{figure}

 Lifar et al. \cite{LifarRFTransformer} and our previous work \cite{JainICADpaper} introduced the RF Transformer and RF Transformer Decoder architectures as RF adaptations of models typically used for natural language tasks. Figure \ref{fig:transformers} shows the RF Transformer model adapted to include the RF FSQ tokenizer. We used the tokenizer layer at the output end of the models to turn the SOI ``tokens'' back into a continuous signal sequence using the upsampler block learned by the RF FSQ tokenizer. While tokenization on the input signal mixture is possible, we hypothesized that it would be more difficult to learn a compressed representation of a signal mixture, especially with high levels of interference present. Additionally, the RF Transformer at inference time employs an autoregressive generation mechanism inspired by large language models (LLMs). In this process, the model generates tokens sequentially, where each token is predicted based on the previously generated tokens and the input conditions. This approach enables the RF Transformer to convert discrete tokens into a continuous signal sequence, one step at a time. Finally, we stacked the real and imaginary components, utilized rotary positional encodings, mean squared error (MSE), and applied random time shifts and phase rotations to simulate transmission impairments \cite{JainICADpaper, LifarRFTransformer}.

\subsection{Training Methodology}\label{method}

\begin{figure*}[htbp]
\centerline{\includegraphics[width=0.8\textwidth]{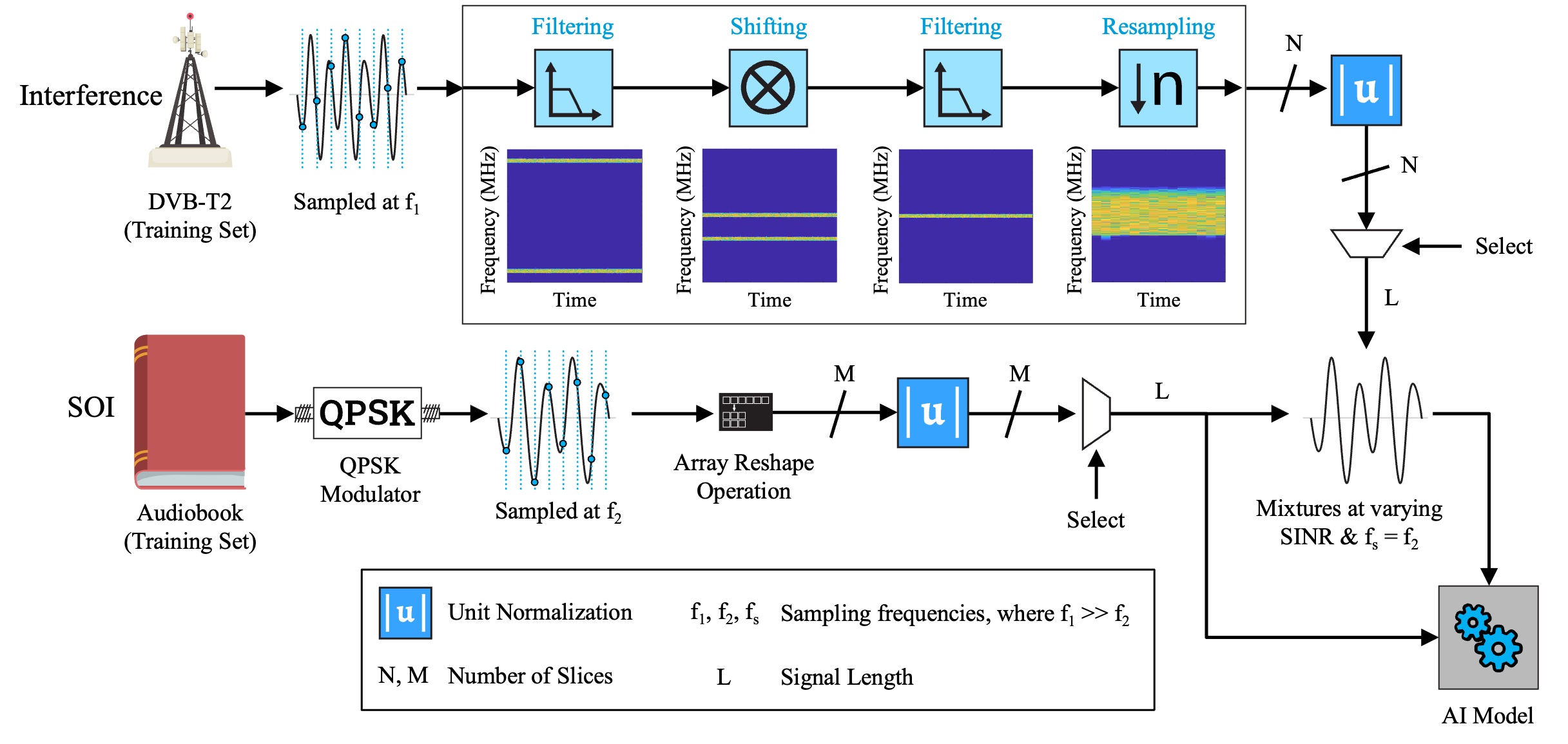}}\caption{ Adapted from \cite{JainICADpaper}. Data preparation process for the training pipeline.}
\label{fig:data_preprocessing}
\end{figure*}

 Our SOI and interference must undergo a number of preprocessing steps before being used in the training process. The sequence of data preprocessing steps shown in Figure \ref{fig:data_preprocessing} follows a similar structure to that described in \cite{JainICADpaper}, with an added bandpass filtering step on the interference data. These combined steps allowed us to bring the SOI and interference datasets to a common sample rate and also carefully control signal magnitudes for stable and efficient learning. The SINR was carefully calculated when mixing the two signal types, spanning a defined range to expose the models to varying levels of interference. These steps are also consistent with the processing that the radio receiver would do. Once this was done, the RF FSQ Tokenizer was trained on the SOI to learn the compressed representation. After this, we trained the RF Transformer model to predict the tokenized representation and then had the tokenizer turn this into a continuous signal representation.

\subsection{Experimental Design: Digital Radio}

 For our experiment, we chose to use a digital radio with a waveform modulation that is commonly used by military personnel, law enforcement officers, and first responders---QPSK (Quadrature Phase Shift Keying) in particular. We constructed mixtures of digitally modulated QPSK voice as the SOI and synthetic digital television, specifically Digital Video Broadcasting$-$Second Generation Terrestrial or DVB-T2 signals, as the interference. The digitally modulated QPSK voice dataset was collected by transmitting 2 hours of an audiobook of Robert Louis Stevenson's \textit{Treasure Island} through a $\mu$-law audio encoder and QPSK modulator and saving the transmitted I/Q signal samples to a file. The audiobook was originally at a 44.1 kHz audio sample rate; however, after performing QPSK modulation with 2 bits per symbol and a sampling rate of 4 samples per symbol, the resulting signal sample rate was 705.6 kHz with a bandwidth of 238.14 kHz. The oversampling factor of 4x was chosen to provide the AI models with repetition to learn from. The DVB-T2 signals had a sample rate of 9.14286 MHz, bandwidth of 8 MHz, and use 64 Quadrature Amplitude Modulation (QAM) as the underlying modulation scheme, though other combinations are possible as well. After applying the preprocessing discussed in section \ref{method}, the data was presented to the models for learning\footnote{ Note that we did not employ k-fold cross-validation, as retraining these large transformer-based models on each fold would be computationally prohibitive; instead a fixed held-out test set was used.}.

\section{Results}

 As in prior work, we evaluated the experiment via the processed SOI quality and overall latency \cite{JainICADpaper}. Models with configurations as described in Table \ref{tab:algorithms} were evaluated.

\begin{table*}[htbp]
\caption{ AI-based Algorithms Architecture Configurations}
\begin{center}
\renewcommand{\arraystretch}{1.2} % Adjust row height for readability
\setlength{\tabcolsep}{3pt} % Reduce column spacing
\begin{tabular}{|c|c|c|c|}
\hline

\textbf{Model Name} & \textbf{RF WaveNet} & \textbf{RF Transformer Decoder} & \textbf{RF Transformer + Tokenizer} \\
\hline
\textbf{Parameter Count} & 3,964,674 & 99,211,808 & 281,102,912 \\
\hline
\textbf{Input/Output Sequence Length} & \multicolumn{3}{c|}{2560 I/Q Samples for both} \\
\hline
\textbf{Tokenizer Sequence Length} & N/A & N/A & 2560 I/Q Samples \\
\hline
\textbf{Number of Layers} & 30 residual & 14 decoder & 14 encoder, 14 decoder \\
\hline
\textbf{Hidden Dimension} & 128 residual channels & \multicolumn{2}{c|}{768} \\
\hline
\textbf{Attention Heads} & N/A & \multicolumn{2}{c|}{12} \\
\hline
\textbf{Receptive Field} & 6139 & \multicolumn{2}{c|}{N/A} \\
\hline
\textbf{Window/Context Size} & N/A & \multicolumn{2}{c|}{16 I/Q Samples for both}\\
\hline

\end{tabular}
\end{center}
\label{tab:algorithms}
\end{table*}

\subsection{SOI Quality: Audio Quality \& Intelligibility}

\begin{figure}[htbp]
\centerline{\includegraphics[width=\columnwidth]{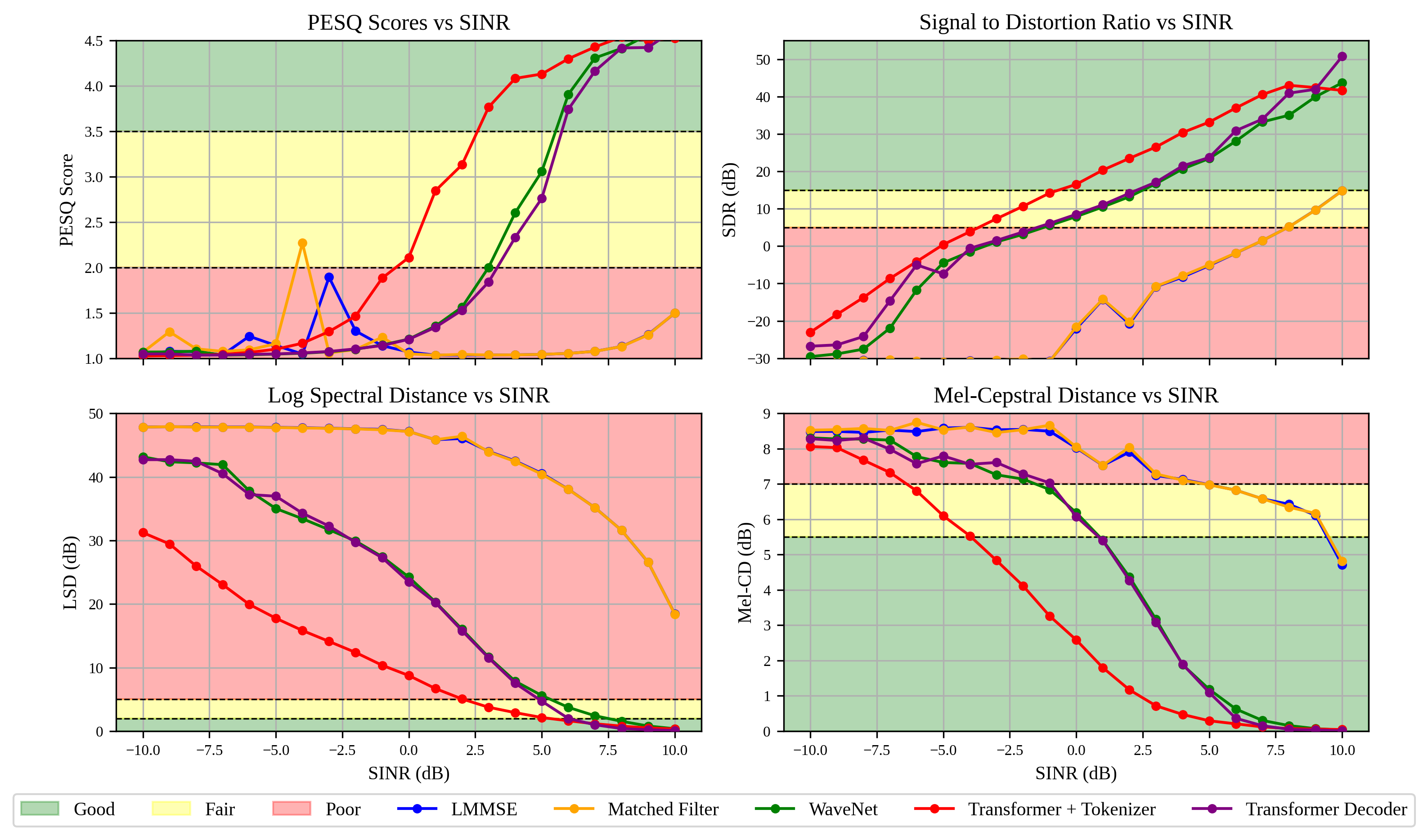}}
\caption{ Audio quality evaluation for digital radio experiment. Metrics include Perceptual Evaluation of Speech Quality (PESQ), Signal to Distortion ratio (SDR), Log-Spectral Distance (LSD), and Mel-Cepstral Distance (Mel-CD).}
\label{fig:audio_quality}
\end{figure}

\begin{figure}[htbp]
\centerline{\includegraphics[width=\columnwidth]{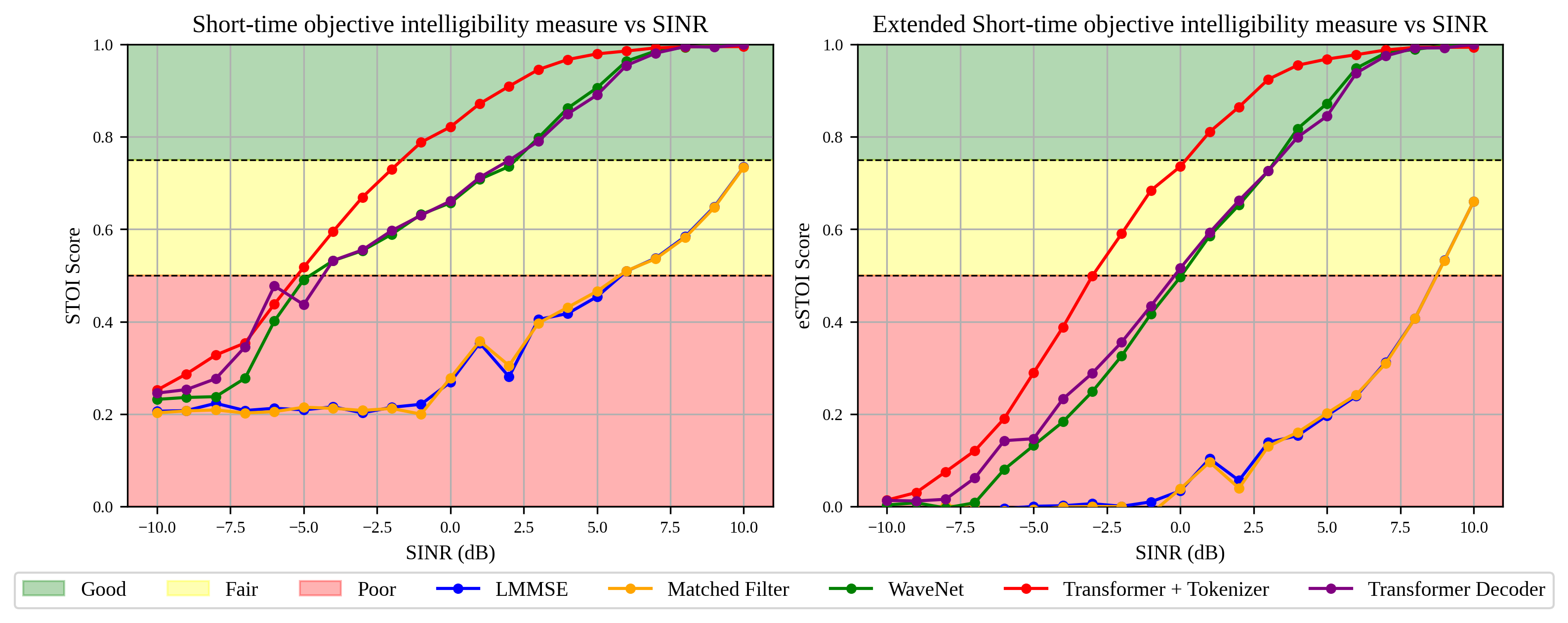}}
\caption{ Audio intelligibility evaluation for digital radio experiment. Metrics include Short-Time Objective Intelligibility (STOI) and extended STOI (eSTOI).}
\label{fig:audio_intelligibility}
\end{figure}

 After processing, the signal converted into audio was compared with the ground truth audio. We adopted similar metrics as \cite{JainICADpaper} including Perceptual Evaluation of Speech Quality (PESQ)\footnote{ The PESQ metric operates on audio with sample rates of 8 or 16 kHz, so for this evaluation our audio was downsampled from 44.1 kHz to 16 kHz.}, Signal to Distortion ratio (SDR), Log-Spectral Distance (LSD), and Mel-Cepstral Distance (Mel-CD)\footnote{ We omit bit error rate (BER) analysis here, as our $\mu$-law audio encoding scales bit-level importance unevenly, making audio-domain metrics more representative of communication quality for this dataset; BER validation is presented separately in \cite{LifarRFTransformer} using a non-communications IQ dataset.}. Figures \ref{fig:audio_quality} and \ref{fig:audio_intelligibility} show the results. As expected, we can see that in all cases the AI-based models outperformed the traditional methods\footnote{ An RF Transformer without a tokenizer and an RF Transformer Decoder with a tokenizer were also trained, but are omitted as these either performed poorly or had inconsistent interference rejection results. We note that the RF Transformer + Tokenizer's larger parameter count relative to the other architectures may also contribute to its performance advantage; isolating the tokenizer's specific contribution independent of model scale is left to future work.}. Additionally, some metrics show the RF Transformer with the tokenizer has good interference rejection for SINR down to -6 dB, which is well beyond the rest and in agreement with other experiments \cite{LifarRFTransformer}. This confirmed our expectations that the addition of the tokenizer was able to boost our performance as shown with the audio metrics. 

\subsection{Overall Latency and Output Throughput}

 As in our earlier work, we focused on the buffer latency and inference time as the two key contributors to overall system latency \cite{JainICADpaper}. Additionally, we considered the output throughput as a design requirement for real-time streaming applications. In this section, the reported latency results used an NVIDIA Jetson AGX Orin edge GPU with the NVIDIA H100 NVL server-grade GPU included as a point of comparison. The equations described in \cite{JainICADpaper} were used to compute these quantities.

 For the digital radio experiment where the sample rate is 705.6 kHz, the buffer latency for a short signal length $L=2560$ is about $4$ msec for a batch size $B=1$. Contrast this with a batch size $B=256$, where the buffer latency is $929$ msec. Waiting too long to fill the buffer with samples for batch processing is unacceptable for a real-time application. For the models reported in Table \ref{tab:algorithms}, we measure the forward pass time per window of signal length or $\tau$ as a function of batch size so that we can compute the inference time. Combining the buffer latency with the inference time will give us an estimate of the overall latency.

\begin{figure}[htbp]
\centerline{\includegraphics[width=\columnwidth]{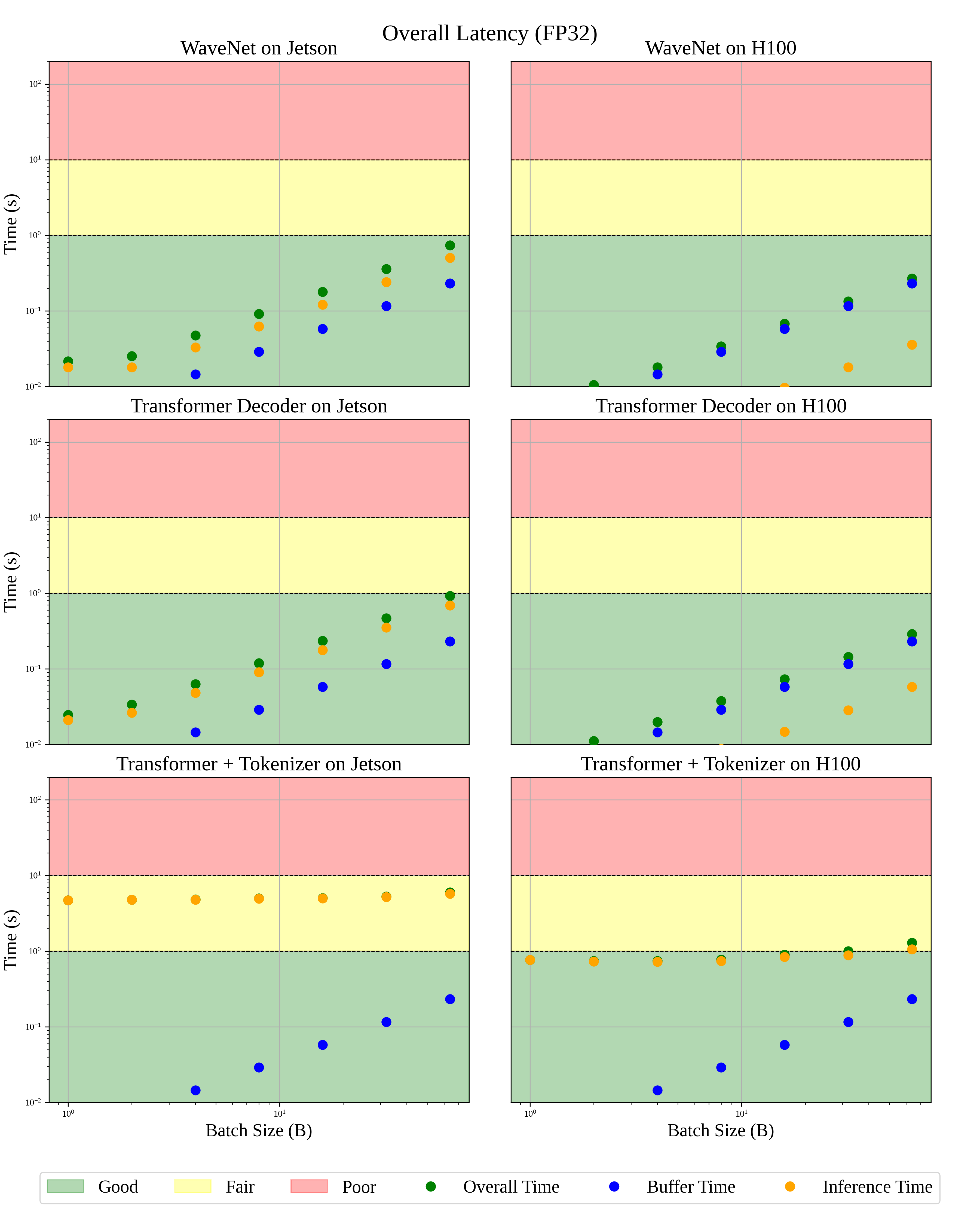}}
\caption{ Overall latencies for the RF WaveNet, RF Transformer Decoder, and RF Transformer + Tokenizer models on the NVIDIA Jetson AGX Orin (left column) and NVIDIA H100 NVL (right column) evaluated for FP32. Note that all three models have good or fair latency for real-time streaming applications on the Jetson.}
\label{fig:latency}
\end{figure}

\begin{figure}[htbp]
\centerline{\includegraphics[width=\columnwidth]{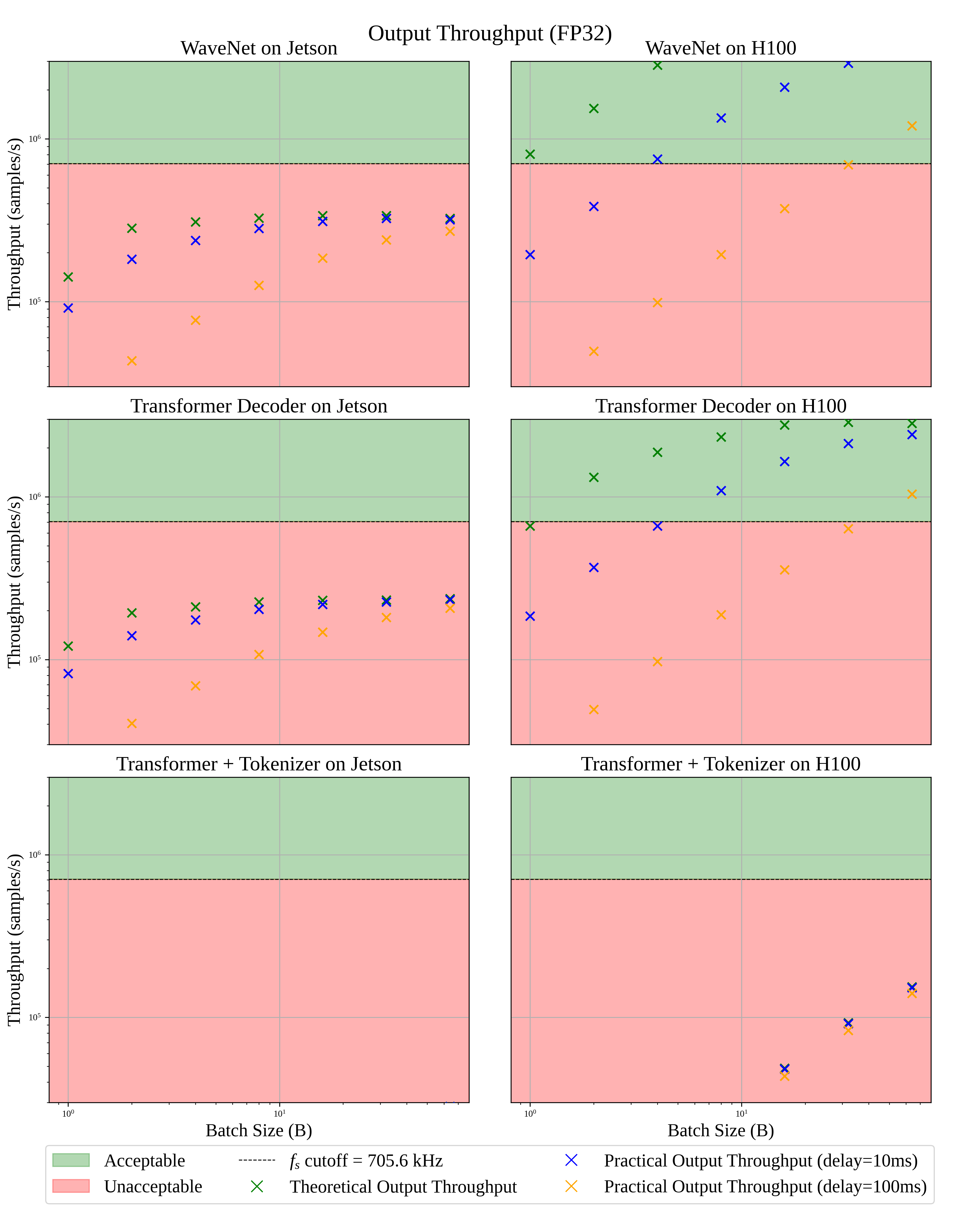}}
\caption{ Output throughput for the RF WaveNet, RF Transformer Decoder, and RF Transformer + Tokenizer models on the NVIDIA Jetson AGX Orin (left column) and NVIDIA H100 NVL (right column) evaluated for FP32. Note that all three models are currently unacceptable for real-time streaming applications on the Jetson, but the Transformer + Tokenizer had very poor throughput in particular thus no points are shown in the range of the plot axes for Jetson.}
\label{fig:throughput}
\end{figure}

 Considering only overall latency, $B=1$ would be the best choice for any of our AI-based methods, as it minimizes buffer latency. Figure \ref{fig:latency} shows that RF WaveNet and the RF Transformer Decoder have similar latencies (less than 1 sec for any $B\leq64$ on the Jetson) while the RF Transformer + Tokenizer's latency is higher and perceptible to an operator, but tolerable within our defined latency budget (around 5 sec for all $B\leq64$ on the Jetson, primarily due to high inference time). However, for applications like real-time voice at the tactical edge where we cannot have a backlog of samples, maximizing output throughput is also a requirement. Figure \ref{fig:throughput} shows that the baseline inference evaluation using floating-point 32-bit (FP32)\footnote{ Even though our models were trained in floating-point 16-bit (FP16), the initial inference evaluation was done at floating-point 32-bit (FP32) as this precision is typically default.} would not allow us to meet this requirement on the Jetson and further optimizations will be required. This is especially important if we consider what the practical output throughput might look like with added miscellaneous delays.

\subsection{Optimizations for Latency and Output Throughput}

 After running the baseline evaluation on our models using FP32, we applied some optimization libraries and techniques to compile and accelerate our models for efficient inference. This should have little to no loss of estimated signal quality.

 To begin, we evaluated the models using the PyTorch library's automatic mixed precision (AMP) capability. This dynamically casts certain operations to lower precision (e.g., floating-point 16-bit (FP16) or brain-floating-point 16-bit (BF16)) during runtime while keeping others in the default FP32 precision. We found on both the H100 and Jetson platforms that this does not offer speedups for our transformer-based models at low batch sizes, but higher batch sizes start to have equal or lower $\tau$ compared to the baseline. On the H100 at FP16 or BF16, $\tau$ was a factor of 0.7-5.8x better or worse depending on the chosen batch size for the RF Transformer Decoder and 0.7-1.9x for the RF Transformer + Tokenizer. On the Jetson, the $\tau$ changed at a factor of 0.8-4.3x for the RF Transformer Decoder and 0.9-1.1x for the RF Transformer + Tokenizer. This is likely the case because transformer-based models rely heavily on operations such as layer normalization, which are typically kept in FP32 precision during AMP to avoid numerical instability. As a result, the dynamic casting overhead introduced by AMP may outweigh the performance benefits of reduced precision for other operations, especially at lower batch sizes where the GPU is not fully utilized. Interestingly, AMP applied to RF WaveNet did not impact $\tau$. This is likely due to its reliance on dilated convolutions, which are memory-bound and less suited for mixed-precision optimizations, and its sequential architecture, which limits GPU parallelism. Overall, using AMP with FP16 gave us acceptable latency and theoretical output throughput for the RF Transformer Decoder model on the Jetson for $B\geq8$, though a practical choice might be $B=16$ or $B=32$ after accounting for miscellaneous processing overheads that come with an implemented system.

\begin{figure}[htbp]
\centerline{\includegraphics[width=\columnwidth]{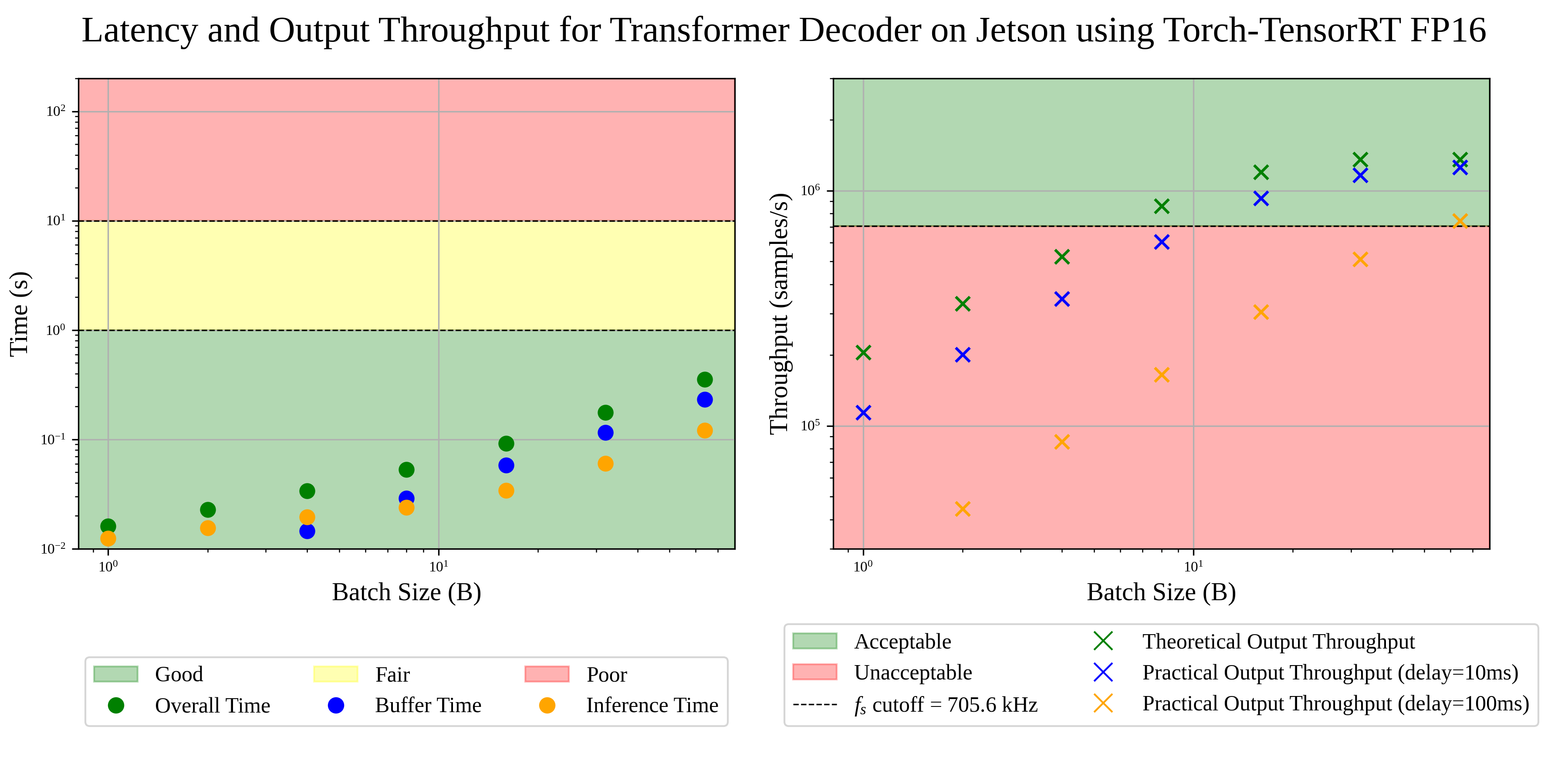}}
\caption{ Optimized latency and output throughput for the RF Transformer Decoder NVIDIA Jetson AGX Orin using the Torch-TensorRT library at FP16 precision.}
\label{fig:optimized_latency_throughput}
\end{figure}

 To improve further, we experimented with the Torch-TensorRT library which applies various optimizations, such as kernel fusion to achieve speedups. We compiled our RF Transformer Decoder with FP32 and FP16 precisions and found an improvement in $\tau$ by a factor of 1.2-5.7x on the Jetson, where FP16 with Torch-TensorRT gave us the best improvement even over the AMP method\footnote{ We were not able to compile the RF Transformer + Tokenizer with the Torch-TensorRT library as there were issues likely related to the dynamic length of the partially decoded SOI discrete tokens that are fed back into the decoder during runtime.}. Figure \ref{fig:optimized_latency_throughput} shows the latency and throughput of our optimized RF Transformer Decoder model using FP16 with the Torch-TensorRT library. Compared to the earlier results in Figures \ref{fig:latency} and \ref{fig:throughput}, we achieve low latency and acceptable output throughput to support a real-time communications application with $B=16$ for example. Here, the buffer time (58 msec) and inference time (34 msec) combine to produce an overall latency of 92 msec, and give us a theoretical output sample throughput of roughly 1.2 MHz, which is higher than the input sample throughput or sample rate of 705.6 kHz.

\section{Discussion}

\subsection{Use Cases}

 Our end goal is to deploy an AI-enabled receiver ``preprocessor'' that can be deployed at the tactical edge. Given the large number of tactical radios that are already in the field, a small receiver front-end could be added in place of existing radio antennas. This avoids requiring design changes to existing radio hardware or software. The ``preprocessor'' concept implies the output of the AI model is converted back to an RF signal which is fed directly into the existing radio connector where the original antenna was installed.

 The experiment described in this work is focused on the case where a real-time voice discussion is occurring. The latency for this type of use case has to be low enough to not be distracting to the operator on either end of the radio interface. The RF Transformer Decoder model can support this while the RF Transformer + Tokenizer would have noticeable delay and also lag behind due to low output throughput. While our prior work utilizes analog FM signals \cite{JainICADpaper}, we experimented with a SOI and interferer that are both digitally modulated, which is closer to what we expect to see in real-world applications. In cases where the communications link includes data, the latency and output throughput requirements may be relaxed depending on the application (e.g., file transmission). In the case of data communications, the RF Transformer + Tokenizer model can also be applied to recover the signal at the receiver when interference is present.

\subsection{Future Work}

 Moving forward, our research will extend to testing our transformer-based signal separation algorithms on new OFDM-based interference sources on which the models have not been trained previously. This will support one of the stated goals of not requiring detailed knowledge of the interferer and still being able to remove it. Digital television signals are convenient to use for future experiments because of the configurability for a range of subcarrier modulation techniques. Additionally, there are a number of other types of digital modulation techniques that we can investigate as our SOI. 

 On the modeling side, we hope to continue to improve estimated signal quality and model robustness, and generalize the models for over-the-air transmissions through better variations of our architecture. We plan to bring the systems to government-sponsored test ranges for further testing in outdoor propagation environments to ensure the same results can be achieved in operational scenarios. Representative datasets will be collected such that generalizable models may be trained across variations in signal/interferer types, geometry, and environmental conditions. Additionally, we aim to improve the latency and output throughput by shortening the required inference processing time through techniques such as post-training quantization and model pruning.

\begin{figure}[htbp]
\centerline{\includegraphics[width=0.7\columnwidth]{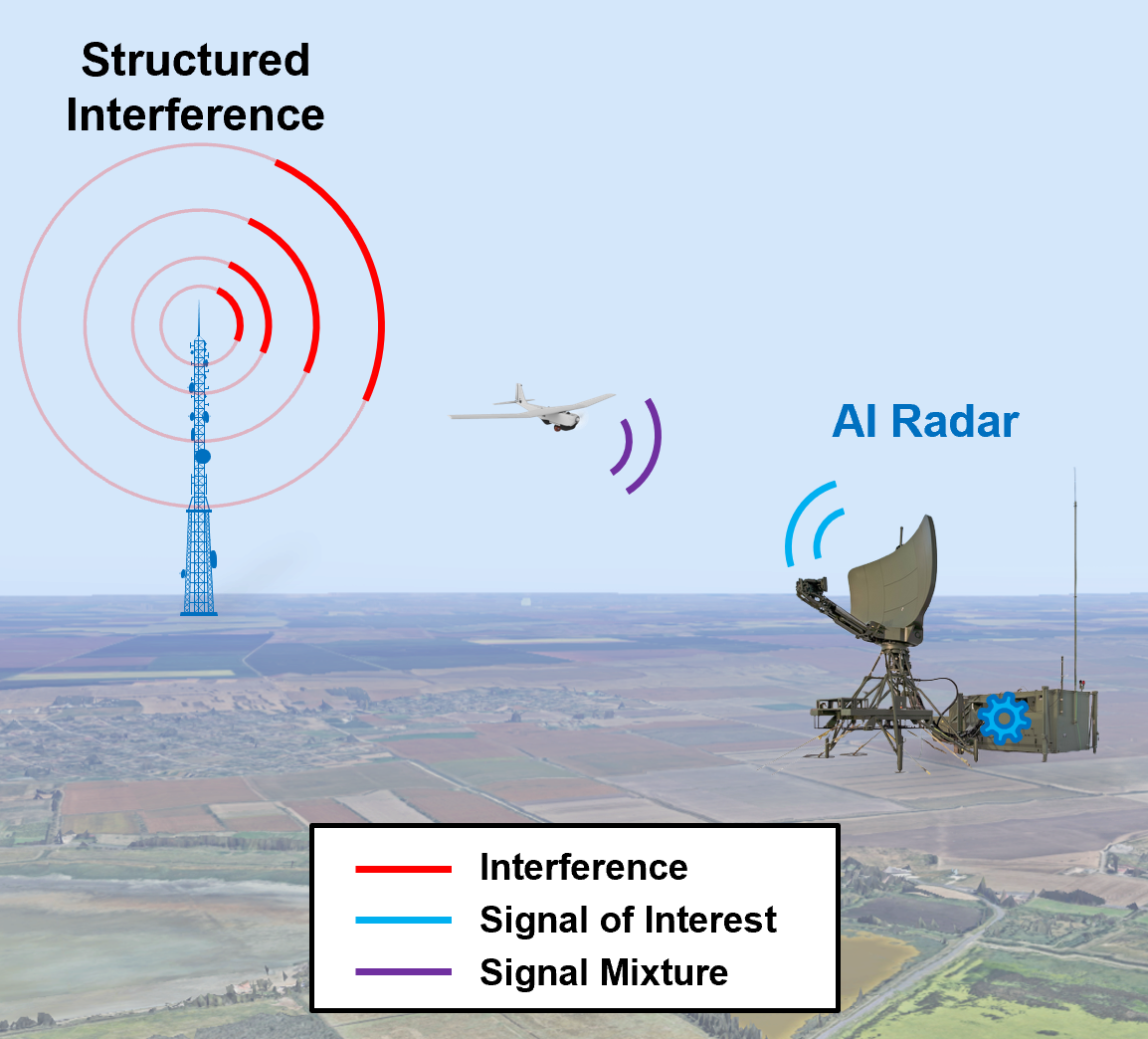}}\caption{ Spectrum sharing use case between radar and a structured interference source.}
\label{fig:use_cases}
\end{figure}

 Finally, we want to expand the types of applications in which these algorithms are used, including multifunction RF systems not necessarily at the tactical edge. In particular, we are concerned with possible interference due to increasing contention between civil spectrum uses and areas traditionally reserved for the military. Spectrum sharing, shown in Figure \ref{fig:use_cases}, would let us use parts of the RF spectrum more efficiently by allowing multiple transmissions (that might normally conflict) to operate in the same band. One such example of this is the proposition to open 5G networks to the S-band part of the spectrum, currently occupied by military radars \cite{breakingdefense}. These radars may become jammed if 5G were allowed to transmit in the band, but AI algorithms such as the ones presented in this paper may be able to offer a way out of this increasingly contentious problem.

\section{Conclusion}

 The results presented demonstrate considerable success toward our goal of operationalizing recent results in AI-enhanced interference rejection to improve the performance of tactical radios. We experimented with a digital SOI and digital interferer and showed that a tokenizer applied with our RF Transformer model can be used to improve upon previous AI-enabled and traditional interference suppression methods. We also demonstrated how higher sample rate datasets may be supported with optimized inference even at the tactical edge for a real-time resilient communications application. We further discussed a number of use cases and future directions for these algorithms including spectrum sharing and data transfer over communications links. In considering both SOI and interferer, AI-enhanced interference rejection in radio transmissions has been shown to improve detection, demodulation, and decoding of signals over a range of SINR levels without having a detailed, design-level knowledge of the interfering signal or the propagation conditions. These same techniques can also be applied to a broader set of military and commercial applications where interference rejection is a crucial issue.

\section*{Acknowledgment}

 The authors acknowledge the MIT Lincoln Laboratory Supercomputing Center for providing HPC resources that have contributed to the research results reported within this paper \cite{8547629}. Additionally, we thank our MIT collaborators and the support given by Major Andrew Xiao and Major Jovan Popovich, without whom this work would not have been possible.

\bibliographystyle{IEEEtran}
\bibliography{references}

\end{document}